\documentclass[conference]{IEEEtran}
\IEEEoverridecommandlockouts
\usepackage{cite}
\usepackage{amsmath,amssymb,amsfonts}
\usepackage{algorithmic,mathcomSTEv4}
\usepackage{graphicx}
\usepackage{textcomp}
\usepackage{xcolor}
\def\BibTeX{{\rm B\kern-.05em{\sc i\kern-.025em b}\kern-.08em
    T\kern-.1667em\lower.7ex\hbox{E}\kern-.125emX}}

\usepackage[]{algorithm2e}

\newtheorem{theorem}{{\textbf{Theorem}}}

\newtheorem{lemma}{{{Lemma}}}
\newtheorem{remark}{{{\textbf{Remark}}}}

\begin{document}

\title{Efficient Federated Learning over Multiple Access Channel with Differential Privacy Constraints\\
}

\author{
\IEEEauthorblockN{Amir Sonee}
\IEEEauthorblockA{
amir.sonee@mail.um.ac.ir}
\and
\IEEEauthorblockN{Stefano Rini}
\IEEEauthorblockA{
\textit{National Chiao Tung University (NCTU), Taiwan} \\
stefano@nctu.edu.tw}

}

\maketitle

\begin{abstract}
In this paper, the problem of  federated learning (FL) through digital communication between clients and a parameter server (PS) over a multiple access channel (MAC), also subject to differential privacy (DP) constraints,  is studied.
More precisely, we consider the setting in which clients in a centralized network are prompted to train a machine learning model using their local datasets.
The information exchange between the clients and the PS takes places over a MAC channel and must also preserve the DP of the local datasets.
%
%
Accordingly, the objective of the clients is to  minimize the training loss subject to (i) rate constraints for reliable communication over the MAC and (ii)  DP constraint over the local datasets.
%
%
For this optimization scenario, we proposed a novel consensus scheme in which digital distributed stochastic gradient descent (D-DSGD) is performed by each client.
To preserve DP,  a digital artificial noise is  also added by the users to the locally quantized gradients.
The performance of the scheme is evaluated in terms of the convergence rate and DP level for a given MAC capacity.
The performance is optimized over the choice  of the quantization levels and the artificial noise parameters.
%
%
Numerical evaluations are presented to validate the performance of the proposed scheme.
\end{abstract}


\section{Introduction}
The ever-increasing need to process large amounts of data generated by users
 networks such as  wireless sensor networks and mobile networks has spurred the development of joint data processing algorithms and network protocols \cite{DL_mobile}.
%
%
As learning is one of the fundamental applications motivating data transfer in such large networks, federated learning (FL) has recently emerged as a promising paradigm for distributed machine learning (ML).
In the FL paradigm, a parameter server (PS) aims at training a global model with the federated collaboration of distributed users, also referred to as local learners.
%
%
The FL model alleviates the need of communicating local data toward the PS by opting to, instead, perform the optimization locally and use consensus over multiple iterations to converge to the optimal solution \cite{Federated_optimization}.
%
%
One way for this paradigm to be implemented, is by having the PS share the current model estimate with the network users at each communication round.
Upon receiving an updated model estimate, the local learners compute the gradients over  the local datasets and update the model estimate using stochastic gradient descent (SGD) method.
The  locally-updated models are then communicated to the PS which merges these estimates to produce an updated central estimate.
%
%
By repeating these two update steps, i.e. the \emph{local} model update and the \emph{global} model update, the clients and the PS can converge to the optimal solution.
%
 %
%
 In practical scenarios, the communication between the PS and the local learners is constrained in either the communication rate,  reliability, or connectivity: for this reason authors have considered variation of the basic FL setting also encompassing such communication constraints  \cite{FL_MAC_DDSGD,FL_wireless_LDP}.
%
The FL model, not only allows a better utilization of computational resources, but also provides some security enhancements.
Since the row data is not communicated to the central server, the users' data is never exposed to the security vulnerability which are intrinsic in especially in wireless communication.
Despite this, it has been shown that local gradients may leak some information about the data endangering the privacy of the model \cite{feature_leakage_FL}.
This possible information leakage  has made the investigation of privacy for the FL model an issue of considerable significance.
Further results including DP analysis of FL model can be found in \cite{cpSGD,DP_client_perspective,FL_DP_analysis}.

\subsubsection*{Contributions}
We provide a theoretical formulation for FL optimization subject to communication and privacy constrains and propose a novel computation and communication scheme that attains learning under  communication and privacy constraints.
%
%
%
%
%
%
%
%
%
The proposed communication scheme generalizes that of \cite{FL_MAC_DDSGD} by
%
adding a digital artificial noise to each client's transmitted message to enhance privacy.
Our achievable scheme also generalizes the one in \cite{cpSGD} by enabling efficient communication over MAC through multi-level stochastic quantization of gradients.
%
Additionally, we show how the
%
convergence rate of this scheme can be optimized  as a function of the quantization levels,  and digital noise parameters for a given set of communication and privacy constraints.
%


\subsubsection*{Notation}
The set of integer numbers between 1 and $n$ is shown by $[n]=\{1,\ldots,n\}$.
With $\|\xv\|_\wp$  we indicate the  $\wp$-norm of vector $\xv$.
%
%
For client $i$ at iteration $t$, $\xv^{t}_{i,j}$ is the $j\textsuperscript{th}$ coordinate of vector $\xv_{i}^{t}=\left[\xv_{i,1}^{t}\ldots\xv_{i,d}^{t}\right]^{\sf{T}}$ and $\Xv^{t}_{i,j}$ is the $j\textsuperscript{th}$ column of matrix $\Xv_{i}^{t}=\left[\Xv^{t}_{i,1}\ldots\Xv^{t}_{i,n}\right]$.
\begin{figure}[htbp]
\centerline{\includegraphics[width=0.5\textwidth]{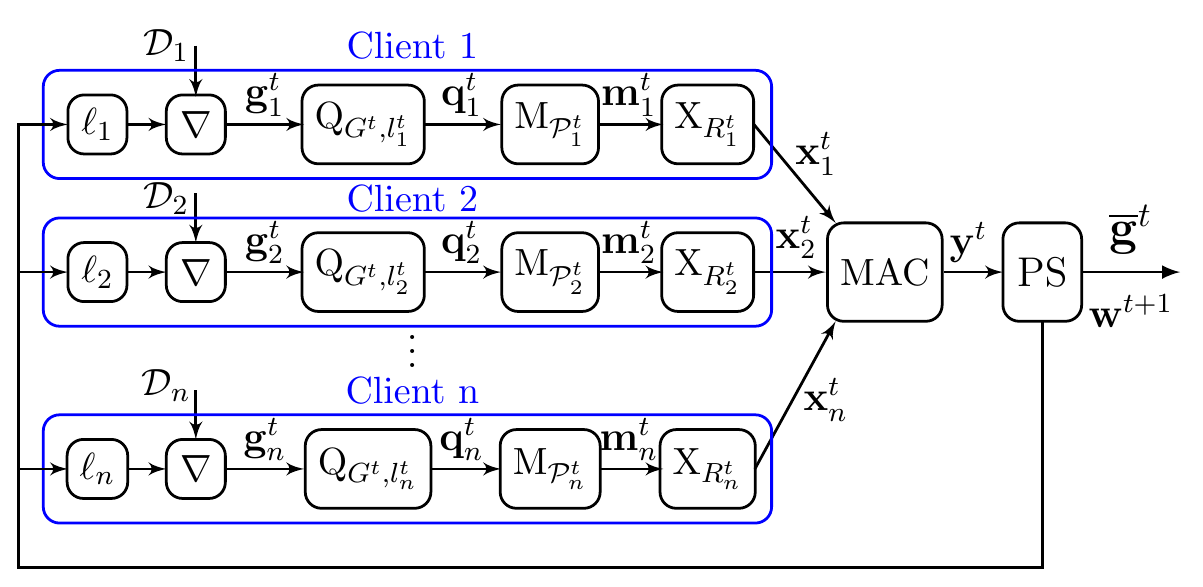}}
\caption{FL over MAC with privacy and efficiency\vspace{-3mm}}
\label{FL_MAC_DDSGD}
\end{figure}
\section{System Model and Problem Formulation}\label{sec:Model and Problem Formulation}
\subsection{FL model with efficiency and privacy}
\label{sec:FL model with efficiency and privacy}
A distributed learning model over a MAC, as introduced in \cite{FL_MAC_DDSGD,FL_wireless_LDP} and shown in Fig.~\ref{FL_MAC_DDSGD}, is characterized by a set $[n]$ of clients communicating with a remote PS over a MAC with the aim of training a machine learning model.
The PS is assumed to have a perfect channel toward each client, separately.
%
Learning can be formulated as minimizing the loss function
\ea{\label{loss_func}
  \bar{\ell}(\wv)\triangleq \dfrac{1}{\left|\Dcal\right|}\sum_{(\dv,v)\in\Dcal}\ell\lb\lb\dv,v\rb,\wv\rb,
}
with respect to the training vector $\wv\in\mathbb{R}^{d}$ during $T$ iterations where
$\ell(\ldotp)$ is the loss function of the learning model, $\Dcal$ is the training database consisting of $|\Dcal|$ data samples $\dv$, constituting columns of $\Dv$, and their corresponding labels $v$.
The solution $\wv^{\ast}=\arg\min_{\wv}\bar{\ell}(\wv)$ for this optimization problem can be carried out collaboratively by the clients and the PS through the iterative use of distributed stochastic gradient descent (DSGD) algorithm. Accordingly, at iteration $t\in[T]$, $t\neq 1$, the parameter vector $\wv^{t}$ is updated by the PS as
\ea{
\wv^{t}=\wv^{t-1}-\gamma^{t-1}\gv^{t-1},
}
where $\gamma^{t-1}$ is the learning rate and $\gv^{t-1}=\gv\lb\wv^{t-1}\rb$ is the stochastic gradient of $\bar{\ell}$ defined as a random function of $\wv^{t-1}$ such that $\mathbb{E}\left[\gv\lb\wv^{t-1}\rb\right]=\nabla\bar{\ell}\lb\wv^{t-1}\rb$ and is computed in a distributed fashion by the clients as $\gv^{t-1}=\sum_{i\in[n]}\gv_{i}^{t-1}/n$ , \cite{FL_Wireless_Fading}. This can be performed by each client upon receiving $\wv^{t-1}$ broadcasted by the PS and through averaging the gradient of the learning loss function over the local datasets as
\begin{equation}\label{local_Gradient}
  \gv_{i}^{t}=\dfrac{1}{\left|\Dcal_i\right|}\sum_{k\in\left[\left|\Dcal_i\right|\right]}
  \nabla\ell\left(\left(\Dv_{i,k},\vv_{i,k}\right),\wv^{t-1}\right).\vspace{-2mm}
\end{equation}
where $\Dcal_{i}=\left\{\left(\Dv_{i,k},\vv_{i,k}\right)\right\}_{k=1}^{|\Dcal_{i}|}$ is the local dataset, as a part of the database $\Dcal$ i.e. $\Dcal=\cup_{i\in[n]}\Dcal_{i}$, used for training at each client $i\in[n]$ with size $|\Dcal_{i}|$ data pairs consisting of a vector data point $\Dv_{i,k}$ which denotes the $k$\textsuperscript{th} data point and its corresponding label, $\vv_{i,k}$, in dataset $\Dcal_{i}$ of client $i$.
%
In order for the above algorithm to guarantee convergence, it is required for the local gradients to have bounded second order moments as $\|\gv_{i}^{t}\|_2\leq D^{t}$.

%
Two constraints that arise naturally in the FL scenario are  (i) a constraint on the communication between the clients and the PS, and (ii) a constraint on the privacy of the local datasets.
These two  additional requirements are mathematically formulated as follows.

\noindent
{\bf {(i)} --  The communication constraint} is concerned with the largest rate of reliable communication that can be attained over the channel as expressed by the MAC capacity.
%
For this constraint to be met, at iteration $t$, each client $i$ applies a digital transmission approach (D-DSGD), as in \cite{cpSGD,FL_MAC_DDSGD}, by quantizing the local gradients into $l_{i}^{t}$ levels with maximum quantized value of $G^{t}$ and through computing a function ${\rm Q}_{G^{t},l_{i}^{t}}:\Rbb^{d}\rightarrow\Rbb^{d}$ of its gradient as $\qv_{i}^{t}={\rm Q}_{G^{t},l_{i}^{t}}\left(\gv_{i}^{t}\right)$.

\noindent
{\bf {(i)} -- The  security constraint}
is concerned with the 
local database privacy which, in turn,  entails the privacy of the local gradients as aggregated at the PS.
For this constraint to be met, one promising way is via the notion of (DP) for the average gradient, as introduced in \cite{DP_book}.
DP is a strong measure of privacy that can ensure privacy for the FL model, even after post-processing performed by the PS.
In order to alleviate the restrictive need to trust the PS, a local DP (LDP) approach is proposed in which a randomized mechanism is applied by each client contributing some randomness to the quantized local gradient $\qv_{i}^{t}$ before transmission toward the PS.
More precisely, this process can be carried out by client $i$ at iteration $t$ via computing a privacy-preserving mechanism ${\rm M}_{\Pcal_{i}^{t}}:\Rbb^{d}\rightarrow \Rbb^{d}$, with the set of parameters $\Pcal_{i}^{t}$, on the quantized gradient as $\mv_{i}^{t}={\rm M}_{\Pcal_{i}^{t}}\left(\qv_{i}^{t}\right)$.
Finally, the perturbed gradient is encoded using a channel coding function of rate $R_{i}^{t}$, ${\rm X}_{R_{i}^{t}}:\Rbb^{d}\rightarrow\Rbb^{N}$, as $\xv_{i}^{t}={\rm X}_{R_{i}^{t}}\lb\mv_{i}^{t}\rb$.

As a result of this procedure, at iteration $t$, client $i$ sends the codeword $\xv_{i}^{t}$ to the PS through $N$ channel uses of the (Gaussian) MAC:  for this model the input/output relationship is expressed as
\ea{
\yv^t=\sum_{i\in[n]}\xv_{i}^{t}+\zv^t,
}
where $\zv^{t}\sim\Ncal\lb 0,\Iv_{N}\rb$ is the additive white Gaussian noise and the channel inputs at each iteration satisfy the average power constraint
\ea{\|\xv_{i}^{t}\|_{2}^{2}\leq NP_i,\ \forall \ i\in[n].}

For error-free decoding of the perturbed gradients at the PS, the rate of transmission by client $i$, $R_{i}^{t}$, should not exceed the capacity of the Gaussian MAC. Accordingly, it is required that
\begin{equation}\label{rate_constraint}
  \sum_{i\in\Scal}R_{i}^{t}\leq\f{1}{2}\log_2\lb 1+\sum_{i\in\Scal}P_i\rb\triangleq C_{\Scal}, \ \ \forall \  \Scal\subseteq [n],
\end{equation}
where $C_{\Scal}$ is the sum-capacity of the MAC, from the subset $\Scal$ of clients to the PS.
Having decoded the perturbed local gradients, the PS next estimates the global gradient by averaging over the decoded gradients as\vspace{-1mm}
\begin{equation}\label{recovered_gradient}
  \gov^{t}=\f{1}{n}\sum_{i \in[n]}\mv_{i}^{t},\vspace{-1mm}
\end{equation}
which is required to release an unbiased estimation of the true global gradient i.e. $\mathbb{E}\lb\gov^{t}\rb=\gv^{t}=\sum_{i \in[n]}\gv_{i}^{t}/n$. The estimation error is evaluated using the mean square error (MSE) as\vspace{-1mm}
\begin{equation}\label{MSE-definition}
  E_t=\max_{\{\Dcal_i\}_{i=1}^{n}}\mathbb{E}\lb\|\gov^{t}-\gv^{t}\|_{2}^{2}\rb.\vspace{-1mm}
\end{equation}

Next, the PS proceeds to update and broadcast the parameter vector $\wv^{t+1}$ to be used for the next iteration by clients over a noiseless link as $\wv^{t+1}=\wv^{t}-\gamma^{t}\gov^{t}$.

We refer this algorithm to as locally differential private with quantized stochastic gradients (LDP-QSGD). The performance of LDP-QSGD is evaluated in terms of the convergence rate and LDP addressing the efficiency and privacy of the algorithm, respectively. The convergence of the FL model is measured using $|\mathbb{E}\lb\ell\lb \wv_{T}\rb\rb-\ell\lb \wv^{\ast}\rb|$ which is expected to vanish as $T$ grows, since the model is learned at the PS and, consequently, at the local learners.
The LDP of the model is measured through the amount of indistinguishability in determining whether a mechanism output is generated from two distinct sample points. For any local gradient pair $\gv_{i}^\prime$ and $\gv_{i}^{\prime\prime}$ and any subset $\Mcal$ of $\sf{Range({\rm M})}$, this can be formally expressed as the probability of $\Mcal$ generated by $\gv_{i}^\prime$ to be within a multiplicative factor of the probability $\Mcal$ generated by $\gv_{i}^{\prime\prime}$:
\ea{\label{LDP_constraint}
  \textsf{Pr}\lb{\rm M}_{\Pcal}\lb{\rm Q}_{D,l}\lb\gv^{\prime}_{i}\rb \rb\!\in\!\Mcal\rb
  \leq e^{\epsilon}\textsf{Pr}\lb{\rm M}_{\Pcal}\lb {\rm Q}_{D,l}\lb \gv_{i}^{\prime\prime}\rb\rb\!\in\!\Mcal\rb\!+\!\delta
}
under which, mechanism ${\rm M}$ is called $\left(\epsilon,\de\right)$ differentially private, \cite{DP_book}.
\subsection{Optimization setting}
For the FL model with digital communication over the MAC subject to privacy and efficiency, as described in Sec. \ref{sec:FL model with efficiency and privacy}, we formulate the following optimization problem.
Using LDP-QSGD, this problem aims at determining the optimum quantization levels $\{l_{i}^t\}_{i\in[n]}$ of the QSGD scheme for local gradients and noise parameters $\{\Pcal_i^t\}_{i\in[n]}$ of the LDP mechanism that maximizes the convergence rate subject to the transmission rates and privacy constraints of clients.
More formally
\begin{IEEEeqnarray}{rcl}\label{Problem_formulation}
  \Pcal:&\min_{\left\{l_{i}^{t},\Pcal_{i}^{t}\right\}_{i\in[n]}}&|\mathbb{E}\lb\bar{\ell}(\wv_{T})\rb-\bar{\ell}\lb \wv^{\ast}\rb|,\nonumber\\
  &\textrm{s.t.}& \epsilon^{t}\leq \Delta_{\epsilon}, \\
  &&\ \sum_{i\in\Scal}R_{i}^{t}\leq C_{\Scal},\ \forall\ \Scal\subseteq[n] \nonumber
\end{IEEEeqnarray}
for $C_{\Scal}$ as in \eqref{rate_constraint}.
%
%
%
It other words, we wish to determine the transmission rates allocation at the clients so that the maximum convergence rate is achieved while also satisfying the privacy and communication constraints.
%
%
The optimization results
allows us to characterize the trade-offs between convergence rate, privacy and capacity.

These questions will be addressed in the remainder of the paper.
\vspace{-2mm}
\subsection{Preliminaries}
\label{sec:Preliminaries}
%
A function $f$ is a $\lambda$-strong convex if $\forall \ \wv,\wv^{\prime}\in\Wcal$ and any subgradient $\gv$ we have  $f\left(\wv^{\prime}\right)\geq f(\wv)+\gv\ldotp\left(\wv^{\prime}-\wv\right)+\frac{\lambda}{2}\left\|\wv^{\prime}-\wv\right\|^2$.
A function $f$ is $\mu$-smooth with respect to the optimum point $\wv^{\ast}$ i.e. $\forall \ \wv\in\mathcal{W}$ we have $f(\wv)-f(\wv^{\ast})\leq\frac{\mu}{2}\left\|\wv-\wv^{\ast}\right\|^2$ which normally appear in least square regressions.
A function $f$ is $L$-Lipschitz continuous if $\forall \ \wv,\wv^{\prime}$ we have $\left\| f\left(\wv^{\prime}\right)-f\left(\wv\right)\right\|\leq L\left|\wv^{\prime}-\wv\right|$, \cite{GD_strong_convex}.
\section{Main Results}
\label{sec:Main Results}
For the problem formulation in Sec. \ref{sec:Model and Problem Formulation}, we provide an upper bound on the convergence rate in \eqref{Problem_formulation}.
To do so, we design a communication and computation scheme referred to as binomial LDP-QSGD (BinLDP-QSGD) which is comprised of a number of elements, each covered in  this section in a separate subsection:
a quantization scheme, in Sec. \ref{sec:Multi-Level QSGD}, and a binomial privacy-preserving mechanism, in Sec. \ref{sec:Binomial privacy-preserving mechanism}.
The performance of the overall scheme is presented in Th. \ref{MSE-DP-Thm} in Sec. \ref{sec:Results for the solution of the optimization problem}. This is specifically investigated in solving a convex learning problem minimizing a convex function $\bar{\ell}$ using unbiased estimates of gradients, \cite{GD_strong_convex}.
More formally, assume that $\Wcal\subseteq\mathbb{R}^{d}$ is a convex set and $\bar{\ell}:\Wcal\rightarrow\mathbb{R}^{d}$ is a differentiable, convex and smooth function. Applying SGD for a finite number of iterations $T$, the goal is to bound the convergence $\bar{\ell}(\wv_T)-\bar{\ell}(\wv^{\ast})$ in the expectation sense assuming that $\bar{\ell}$ achieves its minimum $\wv^{\ast}\in\Wcal$. A standard convergence analysis \cite{GD_strong_convex} assumes three properties for $\bar{\ell}$  as in Sec. \ref{sec:Preliminaries}:
(i) $\bar{\ell}$ is a $\lambda$-strong convex, (ii) $\bar{\ell}$ is $\mu$-smooth with respect to the optimum point $\wv^{\ast}$, and  $\bar{\ell}$ has $L$-Lipschitz continuous gradients.
%

\subsection{Multi-Level QSGD}
\label{sec:Multi-Level QSGD}
When communication among PS and client is subject to a cardinality constraint, one can consider a  variation of SGD in which gradients are quantized before transmission.
Such variation of the classic set up is termed quantized SGD (QSGD):
%
in the proposed scheme,  we consider a multilevel version of the element-wise QSGD that is described as follows for $j$-th element $g=\gv_{i,j}^{t}$ of the gradient vector of client $i$ at iteration $t$. we drop the indices corresponding to these variables for simplicity.
%
%
At each iteration, each client quantizes uniformly and equally each element $g$ of its local gradient vector $\gv$ into $l$ levels such that it lies in $[-G,G]$. This can be done by a simple clipping process so that $-G\leq g\leq G$ where $G$ is sent by the PS to all clients at the beginning of each iteration. The quantized values are spaced with step $s=2G/\lb l-1\rb$. Then, the stochastic procedure assigns the quantized value ${\rm Q}_{G,l}\lb g\rb$ to the element $g$ as
\begin{IEEEeqnarray}{rcl}\label{Multi-level_QSGD}
  {\rm Q }_{G,l}\lb g\rb=\left\{
  \begin{array}{cc}
   B(r+1) & \!\!\!\textrm{w.p.}\ \ \f{g-B(r)}{s} \\
   B(r) & \!\textrm{w.p.}\ \ \f{B(r+1)-g}{s},
  \end{array}\right.
\end{IEEEeqnarray}
where $r\in[0,l)$, $l\geq2$, and $B(r)$ denote respectively the index and the bin to which the element of local gradient belongs. $r$ is such that $g\in\left.\left[B(r),B(r+1)\right.\right)$ and $B(r)$ is given as\vspace{-1mm}
\begin{equation}\label{Bin}
  B(r)\triangleq -G+rs.\vspace{-1mm}
\end{equation}

\subsection{Binomial privacy-preserving mechanism}
\label{sec:Binomial privacy-preserving mechanism}
A simple and widely-used method for preserving privacy is to add Gaussian noise to the local gradients.
While this mechanism is beneficial in practice, it is not suitable when transmitting quantized local gradients.
An alternative privacy mechanism is presented in \cite{cpSGD}: here an i.i.d artificial noise $\nv$ with binomial-distributed elements, having parameters $m$ and $p$ as ${\rm Bin}(m,p)$, is added in a scaled unbiased manner after quantization.
%
%
Accordingly, each client perturbs the quantized vector as:\vspace{-1mm}
\begin{IEEEeqnarray}{rcl}
	\label{transmitted signal}
  \mv&=&{\rm M}_{p,m}\left(\qv\right)
  =\qv+s\lb\nv-mp\rb,
\end{IEEEeqnarray}
and encodes as $\xv={\rm X}_{R}\lb\mv\rb$. As a result of this process, client $i$ transmits the codeword $\xv_{i}^{t}={\rm X}_{R_{i}^{t}}({\rm M}_{p,m_{i}^{t}}({\rm Q}_{G^{t},l_{i}^{t}}(\gv_{i}^{t})))$ at iteration $t$. Therefore, the rate of transmission is $R_{i}^{t}=d\log\left(l_{i}^{t}+m_{i}^{t}\right)/N$ bits per channel use. Hence, for the PS to reliably extract the local gradients, it is required that this amount of information does not exceed the capacity of the MAC. The PS having decoded the transmitted vector $\xv_{i}^{t}$ from $\yv^{t}$, proceeds to compute the average of the local gradients as
\ea{	
	\label{received_average}
  \gov^{t}=\f{1}{n}\sum_{i \in[n]}\xv_{i}^{t}
  =\f{1}{n}\sum_{i \in[n]}\qv_{i}^{t}
  +s_{i}^{t}\lb\nv_{i}^{t}-pm_{i}^{t}\rb.
}

As $\mathbb{E}\left[\qv_{i}^{t}\right]=\gv_{i}^{t}$ and $\mathbb{E}\left[\nv_{i}^{t}\right]=pm_{i}^{t}$, then $\gov^{t}$ is an unbiased estimation of $\gv^{t}$ as desired. Moreover, the performance of the BinLDP-QSGD in terms of the convergence rate and the LDP can be bounded as provided in the next theorem.
\begin{theorem}\label{MSE-DP-Thm}
Assuming the loss function $\bar{\ell}$ is $\lambda$-strong convex and $\mu$-smooth, the convergence rate of the BinLDP-QSGD algorithm with learning rate of $\gamma_{t}=1/(\lambda t)$ can be upper bounded as
  \begin{IEEEeqnarray}{rcl}\label{eq:Convergence-Bound}
    &&|\mathbb{E}\left[\bar{\ell}(\wv_{T})\right]-\bar{\ell}(\wv^{\ast})|\\
    &&\leq\!\f{2\mu}{\lambda^{2}T^{2}}\!\sum_{t=1}^{T}\!\lb\sum_{i\in [n]}\!\f{d\lb G^{t}\rb^2}{n^{2}\lb l_{i}^{t}-1\rb^2}\left[p(1-p)m_{i}^{t}+1/4\right]
    \!+\!(D^{t})^{2}\rb\nonumber
  \end{IEEEeqnarray}
and for any $\delta$, $p$ and $s$ such that $m_{i}^{t}p(1-p)\geq\max\left(23\ln\lb10d/\de\rb,2\Delta_{i,\infty}^{t}/s_{i}^{t}\right)$, $\forall i\in[n]$,
BinLDP-QSGD guarantees an
$(\epsilon^{t},2\de)$ LDP mechanism where
\begin{IEEEeqnarray}{rcl}\label{eq:LDP_result}
\epsilon^{t}&\geq& \f{\Delta_{i,2}^{t}\sqrt{2\ln\f{1.25}{\delta}}}{\sqrt{m_{i}^{t}p(1-p)}}
    +\f{\Delta_{i,2}^{t}c_{p}\sqrt{2\ln\f{10}{\delta}}+\Delta_{i,1}^{t}b_{p}}{m_{i}^{t}p(1-p)\left(1-\f{\delta}{10}\right)} \nonumber\\
     &&+\f{\f{2}{3}\Delta_{i,\infty}^{t}\ln\f{1.25}{\delta}+\Delta_{i,\infty}^{t}d_{p}\ln\f{20d}{\delta}\ln\f{10}{\delta}}
     {m_{i}^{t}p(1-p)},\ \forall i\in[n]
\end{IEEEeqnarray}
and $\Delta_{i,\wp}^{t}$, $\wp\in\{1,2,\infty\}$, are high probability ($\geq 1-\delta$) $\wp$-norm sensitivity bounds as given in \cite{cpSGD},
  \begin{IEEEeqnarray}{rcl}
    \Delta_{i,\infty}^{t} &=& l_{i}^{t}+1 \nonumber\\
    \Delta_{i,1}^{t} &=& \f{\sqrt{d}D^{t}}{G^{t}}(l_{i}^{t}-1)+\sqrt{\f{2\sqrt{d}D^{t}\ln\lb 2/\delta\rb}{G^{t}}(l_{i}^{t}-1)}\!+\!\f{4}{3}\ln\f{2}{\delta}\nonumber\\
    \Delta_{i,2}^{t} &=& \f{D^{t}}{G^{t}}(l_{i}^{t}-1)+\sqrt{\Delta_{i,1}^{t}+\f{2\sqrt{d}D^{t}\ln\lb 2/\delta\rb}{G^{t}}(l_{i}^{t}-1)}.
  \end{IEEEeqnarray}
\end{theorem}
\begin{IEEEproof}
The proof is provided in Appendix. 
\end{IEEEproof}
\begin{remark}
  Under the case of equal quantization levels applied by clients on the local gradients i.e. $l_{i}^{t}=l^{t}$, $\forall i\in[n]$, then the result of Thm. \ref{MSE-DP-Thm} reduces to the one in \cite[\textit{Thm.} 3]{cpSGD}.
\end{remark}

As it can be seen from \eqref{eq:Convergence-Bound}, the maximum convergence rate is achieved for higher levels of quantization and lower levels of artificial noise. However, this is in contrast with the amount of levels for which the maximum privacy is achieved. For this, lower quantization levels and higher noise parameters are required to get small values of LDP in \eqref{eq:LDP_result} resulting in higher privacy. Hence, it is essential to propose an optimal rate allocation scheme based on some performance criteria.
\subsection{Results for the solution of the optimization problem}
\label{sec:Results for the solution of the optimization problem}
The main result of the paper is the following theorem, which characterizes the convergence of the BinLDP-QSGD algorithm.
\begin{theorem}\label{Opt-problem1}
  The optimum values of quantization levels and noise parameters maximizing the convergence rate in Th. \ref{MSE-DP-Thm} are the solution to the following optimization problem
 \begin{IEEEeqnarray}{rcl}\label{Problem_formulation_QSGD_binomial}
  \Pcal_1:&\min_{\left\{l_{i}^{t},m_{i}^{t}\right\}}&\sum_{i \in[n]}\f{d\lb G^{t}\rb^2}{\left(l_{i}^{t}-1\right)^2}\left[  p(1-p)m_{i}^{t}+1/4\right]\nonumber\\
  &\textrm{s.t.}&\f{\Delta_{i,2}^{t}\sqrt{2\ln\f{1.25}{\delta}}}{\sqrt{m_{i}^{t}p(1-p)}}
    +\f{\Delta_{i,2}^{t}c_{p}\sqrt{2\ln\f{10}{\delta}}+\Delta_{i,1}^{t}b_{p}}{m_{i}^{t}p(1-p)\left(1-\f{\delta}{10}\right)} \nonumber\\
    &&+\f{\f{2}{3}\Delta_{i,\infty}^{t}\ln\f{1.25}{\delta}+\Delta_{i,\infty}^{t}d_{p}\ln\f{20d}{\delta}\ln\f{10}{\delta}}{m_{i}^{t}p(1-p)}
    \leq \Delta_{\epsilon},\nonumber\\
  &&m_{i}^{t} p(1-p)\geq\max\!\left\{23\ln\lb 10d/\delta\rb,(l_{i}^{t})^{2}-1/G^{t}\right\}\!,\nonumber\\
  &&\sum_{i\in\Scal}d\log\left(m_{i}^{t}+l_{i}^{t}\right)\leq NC_{\Scal},\ \Scal\subseteq[n],\nonumber\\
  &&m_{i}^{t},l_{i}^{t}\in \mathbb{Z}^{+},\ l_{i}^{t}\geq 2,\ \forall \  i\in[n].
\end{IEEEeqnarray}
\end{theorem}

The optimization problem stated above is an instance of constrained integer non-linear  programming (INLP) . Since objective function is non-convex, the common method of relaxing the integer constraints and rounding the solution of the resulting problem as in \cite{FL_MAC_DDSGD} is not working in this case. Hence, it is required to go through some numerical algorithm to deal with the solution of this problem.
One main result of this theorem indicates that in order to maximize the convergence rate for a given level of privacy and capacity limit, it is required to allocate the highest level of quantization and lowest level of artificial noise allowed by the capacity limit that satisfies the LDP constraint for almost equality i.e. $\epsilon^{t}=\De_\epsilon$.

\section{Simulation Results}
\label{sec:Numerical Results}
\begin{figure}[htbp]
\centerline{\includegraphics[width=0.45\textwidth]{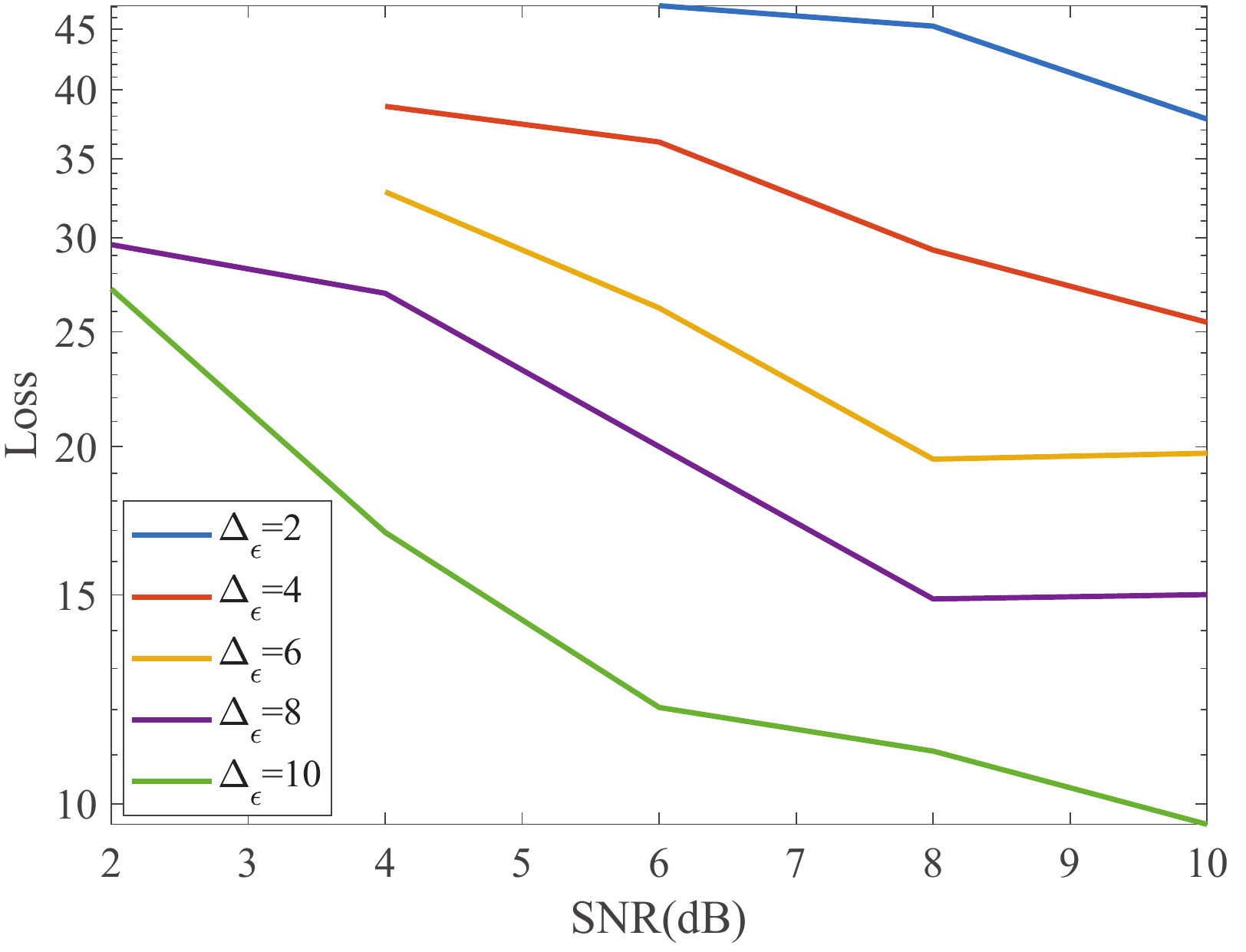}}
\vspace{-3mm}
\caption{\vspace{-4mm}Loss vs $\sf{SNR}$ threshold for $T=100$ and different DP thresholds}
\vspace{-3mm}
\label{fig:loss-snr}
\end{figure}

To evaluate the performance of the FL model under the BinLDP-QSGD algorithm, we consider the loss function of the learning model to be the ridge regression function as
\ea{
\ell\lb(\dv,v),\wv\rb=\lb\wv^{\sf{T}}\dv-v\rb^{2}+\f{\beta}{2}\|\wv\|^{2}.
}
where $\beta=10^{-3}$ is the regularization parameter and the database consists of $|\Dcal|=10^4$ i.i.d sample data points $(\dv,v)$ of dimension $d=10$ generated according to $\Ncal(0,\Iv_{d+1})$. The learning algorithm is performed by $K=2$ users each having a dataset of size $|\Dcal_{1}|=|\Dcal_{2}|=5000$ communicating with the PS through $N=40$ channel uses and equal average power of $P_{1}=P_{2}=10$. Also, local gradient and quantization clipping parameters are set to $G^{t}=D^{t}=4$ and the output of the mechanism is intended to achieve $\lb\De_{\epsilon},\de\rb=\lb 4,10^{-2}\rb$ LDP using binomial mechanism of $p=1/2$.
\begin{figure}[htbp]
\centerline{\includegraphics[width=0.45\textwidth]{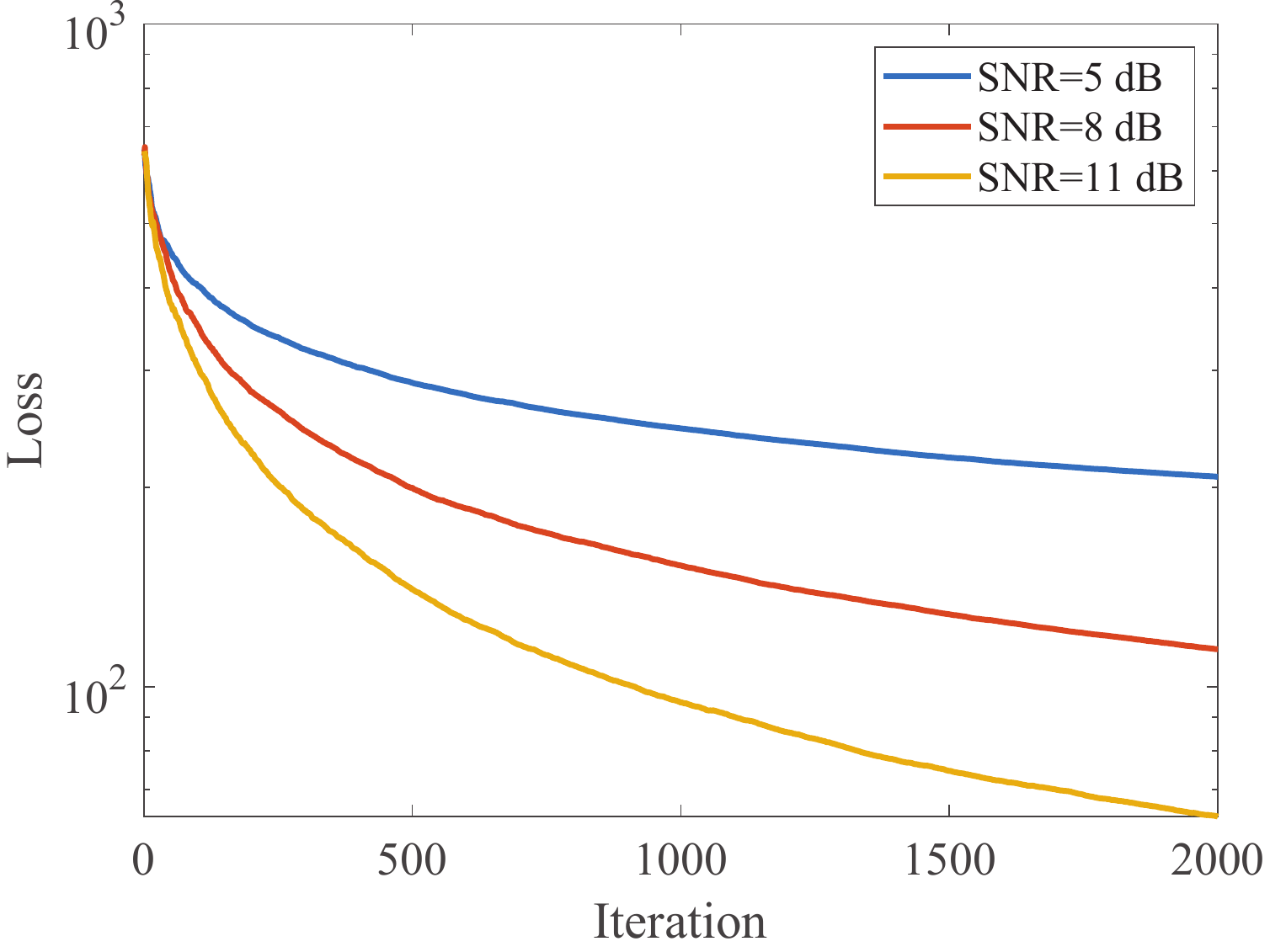}}
\vspace{-0.35 cm}
\caption{\vspace{-4mm}Loss vs number of iterations for $\epsilon=4$ and different values of \sf{SNR}}
\label{fig:loss_iteration_snr}
\end{figure}
\begin{figure}[htbp]
\centerline{\includegraphics[width=0.45\textwidth]{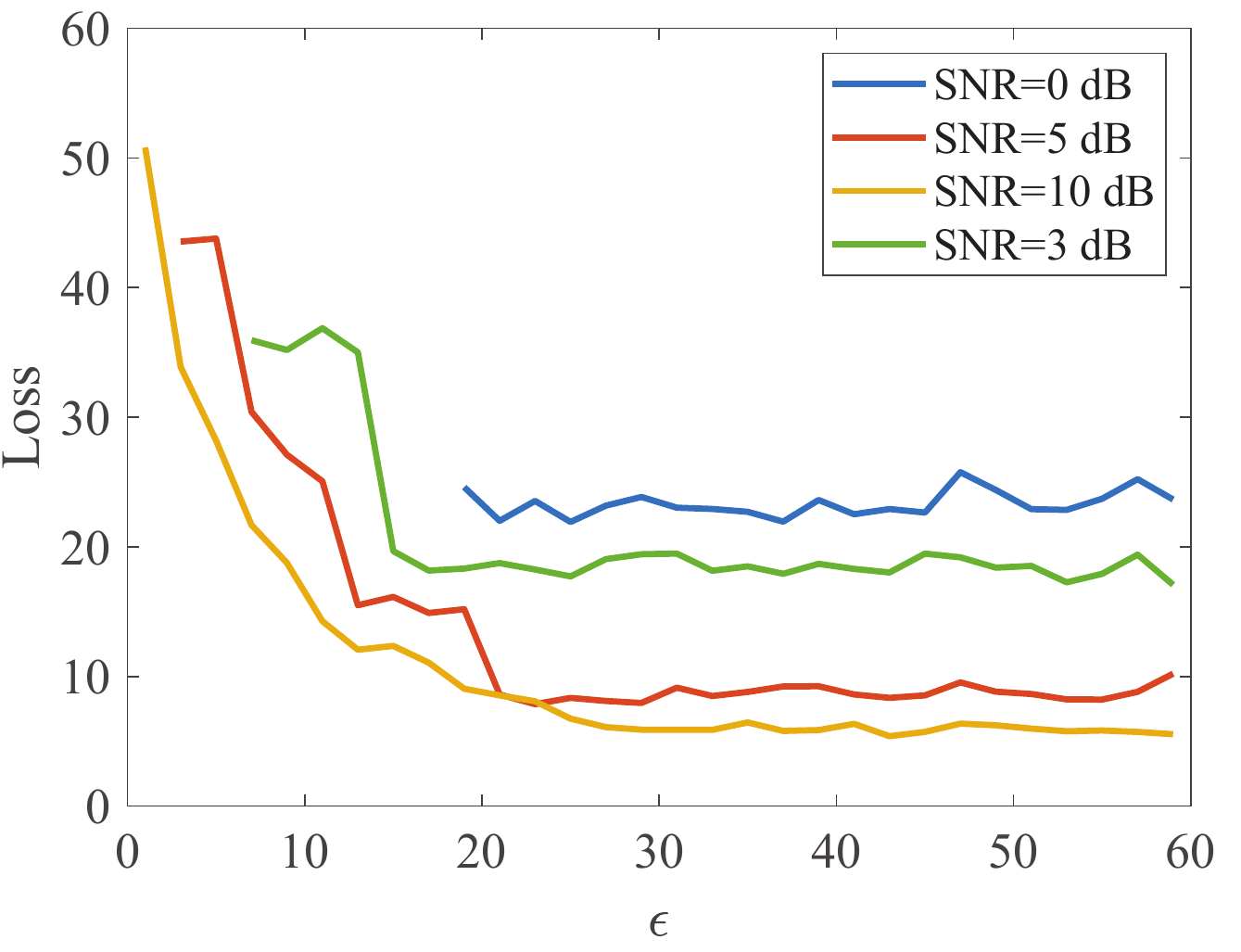}}
\vspace{-3mm}
\caption{\vspace{-4mm}Loss vs privacy threshold for $T=100$ and different values of \sf{SNR}}
\vspace{-3mm}
\label{fig:loss-DP}
\end{figure}
The impact of $\sf{SNR}$ on the convergence rate of the BinLDP-QSGD algorithm is shown in Fig.~\ref{fig:loss_iteration_snr}. As the $\sf{SNR}$ values increase, the training loss decreases with the number of iterations and hence accuracy increases. Fig.~\ref{fig:loss-DP} shows the convergence rate of the training algorithm in terms of the threshold LDP. As the minimum level of required privacy is increased, the training loss decreases as a result of the higher level of quantization level and lower noise parameter that can be achieved for a given fixed channel capacity as the privacy level decreases. Moreover, as the $\sf{SNR}$ value increases, it is possible to achieve higher level of privacy for a fixed loss. The same trend can be seen when we discuss in terms of the $\sf{SNR}$ values as shown in Fig.~\ref{fig:loss-snr} where higher levels of privacy require more $\sf{SNR}$ values while having lower convergence rate and as the privacy measure decreases, less power is needed to reach the same loss.
\section{Conclusion}
\label{sec:Conclusion}
In this paper, the problem of efficient and private communication over MAC in FL model was considered.
In this model, each client sends a digital version of the local gradient using QSGD with specific level of quantization and meanwhile adds binomial noise of discrete values to these quantized gradients in order to preserve the expected level of DP. For this FL model, the differential privacy and convergence rate were derived. The problem was formulated as to maximize the convergence rate subject to capacity and privacy constraints for which an analytic approach to the optimal solution was presented.
 It was shown that to meet both efficiency and privacy in FL model over MAC while having faster convergence to the optimum value, we need to work at lower transmission rates compared with the case where just the efficiency is considered.

\bibliographystyle{IEEEtran.bst}
\bibliography{FL-ref}
\newpage
\section*{Appendix}
\label{App:proof of Thm1}
To prove this, a common result presented by \cite{GD_strong_convex} is used. This result states that using SGD method of learning rate $\gamma_{t}=1/(\lambda t)$ for optimization of a given $\lambda$-strong convex and $\mu$-smooth loss function $\bar{\ell}(\wv^{\ast})$, the convergence of the algorithm can be bounded as below provided that the estimated gradient is stochastically unbiased and has a bounded second-order moment $\mathbb{E}\left[\left\|\gov_{t}\right\|_{2}^{2}\right]\leq \sigma_{g,t}^{2}$, \cite{FL_MAC_DDSGD}:\vspace{-2mm}
\begin{equation}\label{eq:Convergence-bound-convex-loss}
  |\mathbb{E}\left[\bar{\ell}(\wv_{T})\right]-\bar{\ell}(\wv^{\ast})|\leq \f{2\mu}{\lambda^{2}T^{2}}\sum_{t=1}^{T}\sigma_{g,t}^{2}.\vspace{-2mm}
\end{equation}
The second-order moment $\sigma_{g,t}^{2}$ can be bounded as
\begin{IEEEeqnarray}{rcl}\label{eq:second-order-moment-estimation}
   \sigma_{g,t}^{2} &=& \mathbb{E}\left[\left\|\gov_{t}\right\|_{2}^{2}\right]
   =\mathbb{E}\left[\left\|\gov_{t}-\gv_{t}\right\|_{2}^{2}\right]+\left\|\gv_{t}\right\|_{2}^{2}\nonumber\\
   &\stackrel{(a)}\leq&\mathbb{E}\left[\left\|\gov_{t}-\gv_{t}\right\|_{2}^{2}\right]+(D^{t})^{2},
\end{IEEEeqnarray}
where $(a)$ follows since
the local gradients are bounded as $\left\|\gv_{i}^{t}\right\|_{2}\leq D^{t}$, $\forall i\in\Ncal$, satisfying the Lipschitz condition.

The variance of the global gradient estimation (MSE) follows from the definition and taking into account \eqref{received_average} as
  \begin{IEEEeqnarray}{rcl}\label{eq:variance_estimation}
  \!\!&\mathbb{E}&\left[\left\|\gov_{t}-\gv_{t}\right\|_{2}^{2}\right]\nonumber\\ \!\!&=&\mathbb{E}\!\left[\left\|\f{1}{n}\sum_{i \in[n]}\lb\qv_{i}^{t}-\gv_{i}^{t}\rb
  +\f{1}{n}\sum_{i \in [n]}s_{i}^{t}\ldotp\left(\zv_{i}^{t}-m_{i}^{t}p\right)\!\right\|_{2}^{2}\right] \nonumber\\
  &\leq& \f{1}{n^2}\sum_{i\in [n]}\mathbb{E}\left[\left\|\qv_{i}^{t}-\gv_{i}^{t}\right\|_{2}^{2}\right]
  +\f{1}{n^2}\sum_{i\in [n]}\left(s_{i}^{t}\right)^{2}\mathbb{E}\left[\left\|\zv_{i}^{t}-m_{i}^{t}p\right\|_{2}^{2}\right]\nonumber\\
  &=& \f{1}{n^2}\sum_{i\in [n]}\sum_{j=1}^{d}\mathbb{E}\left[\left(\qv_{i,j}^{t}-\gv_{i,j}^{t}\right)^{2}\right]\nonumber\\
  &&+\f{1}{n^2}\sum_{i \in [n]}\left(s_{i}^{t}\right)^2\sum_{j=1}^{d}\mathbb{E}\left[\left(\zv_{i,j}^{t}-m_{i}^{t}p\right)^{2}\right]\nonumber\\
  &\stackrel{(b)}\leq&\f{d}{4n^2}\sum_{i \in [n]}\left(s_{i}^{t}\right)^2+\f{d}{n^2}\sum_{i \in [n]}\left(s_{i}^{t}\right)^{2}\ldotp m_{i}^{t}p(1-p)\nonumber\\
  &=&\f{d}{n^2}\sum_{i\in[n]}\left(s_{i}^{t}\right)^2\left[\f{1}{4}+m_{i}^{t}p(1-p)\right].
\end{IEEEeqnarray}
where $(b)$ follows by $\Ebb((\qv_{i,j}^{t}-\gv_{i,j}^{t})^{2})\leq (s_{i}^{t})^{2}$. As a result, \eqref{eq:Convergence-Bound} is concluded by \eqref{eq:Convergence-bound-convex-loss}, \eqref{eq:second-order-moment-estimation} and \eqref{eq:variance_estimation}.

Before addressing the privacy issue of the BinLDP-QSGD algorithm, it is required to provide the Bernstein’s inequality used in determining the privacy level of the mechanism.
\begin{lemma}[Bernstein’s inequality]
  Assume that $X_1,X_2,\dots,X_n$ are $n$ independent random variables of zero mean and finite variance i.e. $\Ebb(X_i)=0$ and $\mathbb{V}{\rm AR}(X_i)=\sgs_{i}$ having $|X_i|\leq M$ with probability 1, then for any $\de>0$, 
  \ea{
  \textsf{Pr}\lb\sum_{i\in [n]}X_{i}\geq\sqrt{2\sum_{i\in [n]}\sgs_{i}\ln\lb\f{1}{\de}\rb}+\f{2}{3}M\ln\lb\f{1}{\de}\rb\rb\leq\de.
  }
\end{lemma}

For the LDP analysis, it is sufficient to show that the composition of the quantization and binomial mechanism is $(\epsilon,\de)$ LDP which is equivalent to show that $\forall\ i\in[n]$,
\ea{
\textsf{Pr}\lb\ln\lb\f{\textsf{Pr}\lb\textrm{M}_{p,m_{i}^{t}}(\textrm{Q}_{G,l_{i}^{t}}(\gv_{i}^{\prime t}))=\yv\rb}{\textsf{Pr}\lb\textrm{M}_{p,m_{i}^{t}}(\textrm{Q}_{G,l_{i}^{t}}(\gv_{i}^{t}))=\yv\rb}\rb\leq\epsilon\rb\geq 1-\de.
}
Accordingly, the privacy loss can be bounded as
\begin{IEEEeqnarray}{rcl}
\epsilon_i^{t}&=&\ln\lb\f{\textsf{Pr}\lb\mv_{i}^{\prime t}=\yv\rb}{\textsf{Pr}\lb\mv_{i}^{t}=\yv\rb}\rb\nonumber\\
&=&\ln\f{\textsf{Pr}\!\lb\lb\qv_{i}^{\prime t}-\qv_{i}^{t}\rb+\qv_{i}^{t}+s_{i}^{t}\lb\zv_{i}^{t}-m_{i}^{t}p\rb=\yv\rb}{\textsf{Pr}\lb \qv_{i}^{t}+s_i^{t}\lb\zv_{i}^{t}-m_{i}^{t}p\rb=\yv\rb}\nonumber\\
&=&\ln\f{\textsf{Pr}\lb\zv_{i}^{t}=(\yv-\qv_{i}^{t})/s_{i}^{t}+m_{i}^{t}p-\lb\qv_{i}^{\prime t}-\qv_{i}^{t}\rb/s_{i}^{t}\rb}{\textsf{Pr}\lb \zv_{i}^{t}=(\yv-\qv_{i}^{t})/s_{i}^{t}+m_{i}^{t}p\rb}\nonumber\\
&=&\ln\f{\textsf{Pr}\lb\zv^{t}_{i}=\tilde{\yv}_{i}^{t}-\tilde{\qv}_{i}^{t}/s_{i}^{t}\rb}{\textsf{Pr}\lb\zv^{t}_{i}=\tilde{\yv}_{i}^{t}\rb}\nonumber\\
&=&\ln\prod_{j\in[d]}\f{\textsf{Pr}\lb\zv^{t}_{i,j}=\tilde{\yv}_{i,j}^{t}-\tilde{\qv}^{t}_{i,j}/s_{i}^{t}\rb}{\textsf{Pr}\lb \zv^{t}_{j}=\tilde{\yv}^{t}_{i,j}\rb}\nonumber\\
&=&\sum_{j\in[d]}\ln\lb\f{\textsf{Pr}\lb\zv^{t}_{i,j}=\tilde{\yv}^{t}_{i,j}-\tilde{\qv}^{t}_{i,j}/s_{i}^{t}\rb}{\textsf{Pr}\lb \zv^{t}_{i,j}=\tilde{\yv}^{t}_{i,j}\rb}\rb\nonumber\\
&\stackrel{(c)}\leq&\sum_{j\in[d]}\ln\lb\exp\lb\f{\tilde{\qv}^{t}_{i,j}}{s_{i}^{t}}\ln\lb\f{\lb\tilde{\yv}^{t}_{i,j}+1\rb(1-p)}{p\lb m_{i}^{t}-\tilde{\yv}^{t}_{i,j}+1\rb}\rb\rb\rb\nonumber\\
&=&\sum_{j\in[d]}\f{\tilde{\qv}^{t}_{i,j}}{s_{i}^{t}}\ln\lb \f{\lb\tilde{\yv}^{t}_{i,j}+1\rb(1-p)}{p\lb m_{i}^{t}-\tilde{\yv}^{t}_{i,j}+1\rb}\rb\nonumber\\
&\stackrel{(d)}=&\sum_{i\in[n]}\f{\tilde{\qv}^{t}_{i,j}\lb\tilde{\yv}^{t}_{i,j}-m_{i}^{t}p\rb}{s_{i}^{t}m_{i}^{t}p(1-p)}\!
+\sum_{j\in[d]}\f{\tilde{\qv}^{t}_{i,j}}{s_{i}^{t}}\ldotp\f{1-2p}{m_{i}^{t}p(1-p)}\nonumber\\
&&\!+\!\sum_{j\in[d]}\!\!\f{\tilde{\qv}^{t}_{i,j}}{s_{i}^{t}}\!\!\lb\!\ln\!\lb\!\f{\lb\tilde{\yv}^{t}_{i,j}\!+\!1\rb\!(1\!-\!p)}{p\!\lb m_{i}^{t}\!-\!\tilde{\yv}^{t}_{i,j}+1\rb}\rb\!-\!\f{\tilde{\yv}^{t}_{i,j}\!+\!1}{m_{i}^{t}p}\!+\!\f{m_{i}^{t}\!-\!\tilde{\yv}^{t}_{i,j}\!+\!1}{m_{i}^{t}(1-p)}\!\rb
\nonumber\\
&\stackrel{(e)}\leq&\f{\|\tilde{\qv}_{i}^{t}\|_{2}\sqrt{2\ln(1.25/\de)}}{s_{i}^{t}\sqrt{m_{i}^{t}p(1-p)}}
+\f{\f{2}{3}\|\tilde{\qv}_{i}^{t}\|_{\infty}\ln(1.25/\de)}{s_{i}^{t}m_{i}^{t}p(1-p)}\nonumber\\
&&+\f{\|\tilde{\qv}_{i}^{t}\|_{1}(1-2p)}{s_{i}^{t}m_{i}^{t}p(1-p)}
+\f{\f{2}{3}\|\tilde{\qv}_{i,j}^{t}\|_{1}(p^2+(1-p)^2)}{s_{i}^{t}m_{i}^{t}p(1-p)(1-\de/10)}\nonumber\\
&&+\f{\|\tilde{\qv}_{i}^{t}\|_{2}c_{p}\sqrt{\ln(10/\de)}}{s_{i}^{t}m_{i}^{t}p(1-p)\sqrt{1-\de/10}}\nonumber\\
&&+\f{\f{4}{3}(p^{2}+(1-p)^{2})\|\tilde{\qv}^{t}_{i,j}\|_{\infty}\ln(\f{10}{\de})\ln(\f{20d}{\de})}{s_{i}^{t}m_{i}^{t}p(1-p)}
\end{IEEEeqnarray}
where $(c)$ follows by \cite[\textit{Lemma 5}]{cpSGD}, $(e)$ follows by applying the Bernstein’s inequality to the first summation in $(d)$ holding with probability greater than $1-\de/1.25$ and also using \cite[\textit{Lemma 6}]{cpSGD} for the third summation in $(d)$ holding with probability greater than $1-2\de/10$. As a result, the LDP result holds with probability greater than $1-\de$.
\end{document}